\documentclass[letterpaper]{article} 
\usepackage{aaai23}  
\usepackage{times}  
\usepackage{helvet}  
\usepackage{courier}  
\usepackage[hyphens]{url}  
\usepackage{graphicx} 
\usepackage{natbib}  
\usepackage{caption} 
\usepackage{algorithm}
\usepackage{algorithmic}

\usepackage{newfloat}
\usepackage{listings}
\DeclareCaptionStyle{ruled}{labelfont=normalfont,labelsep=colon,strut=off} 
\floatstyle{ruled}
\newfloat{listing}{tb}{lst}{}
\floatname{listing}{Listing}
\usepackage{booktabs}
\usepackage{amsfonts}
\usepackage{nicefrac}
\usepackage{microtype}
\usepackage{multirow}
\usepackage{verbatim}
\usepackage{enumitem}
\usepackage{bm}
\usepackage{amsmath}
\usepackage{amssymb}
\usepackage[british,american]{babel}
\usepackage[table]{xcolor}
\usepackage{arydshln}
\usepackage{hyperref}
\hypersetup{bookmarksopen=true, bookmarksopenlevel=2, bookmarksnumbered=true}
\usepackage{makecell}
\usepackage{tabulary}
\definecolor{demphcolor}{RGB}{144,144,144}
\newcommand{\demph}[1]{{#1}}
\definecolor{mygray}{gray}{0.4}
\usepackage{pifont}
\newcommand{\cmark}{\ding{51}}%
\newcommand{\xmark}{\ding{55}}%

\newlength\savewidth\newcommand\shline{\noalign{\global\savewidth\arrayrulewidth
  \global\arrayrulewidth 1pt}\hline\noalign{\global\arrayrulewidth\savewidth}}

\newcommand{\tablestyle}[2]{\setlength{\tabcolsep}{#1}\renewcommand{\arraystretch}{#2}\centering\footnotesize}

\definecolor{citecolor}{HTML}{0071BC}
\definecolor{linkcolor}{HTML}{ED1C24}

\newcommand{\vqa}[1]{VQAv2}

\newcommand{\methodname}[0]{\textsc{BridgeTower}}
\newcommand{\modelbase}[0]{$_{\text{BASE}}$}
\newcommand{\modellarge}[0]{$_{\text{LARGE}}$}
\newcommand{\modelhuge}[0]{$_{\text{HUGE}}$}
\newcommand{\metername}[0]{\textsc{Meter}}
\newcommand{\vlmoname}[0]{\textsc{VLMo}}
\newcommand{\twotower}[0]{\textsc{Two-Tower}}

\newcommand{\coattention}[0]{\textit{co-attention}}

\newcommand{\eg}[0]{\textit{e.g.},}
\newcommand{\ie}[0]{\textit{i.e.},}

\definecolor{gain}{HTML}{34a853}  %

\definecolor{lost}{HTML}{ea4335}  %

\definecolor{baselinecolor}{gray}{.9}

\title{
    {BridgeTower}: Building Bridges Between Encoders \\ in Vision-Language Representation Learning
}

\author {
    Xiao Xu\textsuperscript{\rm 1, 2}\thanks{\ Contribution during internship at Microsoft. $^\dagger$\ Contact Person},
    Chenfei Wu\textsuperscript{\rm 2},
    Shachar Rosenman\textsuperscript{\rm 3},
    Vasudev Lal\textsuperscript{\rm 3},
    Wanxiang Che\textsuperscript{\rm 1$^\dagger$},
    Nan Duan\textsuperscript{\rm 2$^\dagger$}
}

\affiliations {
    \textsuperscript{\rm 1}Research Center for Social Computing and Information Retrieval, Harbin Institute of Technology \\
    \textsuperscript{\rm 2}Microsoft Research Asia, \textsuperscript{\rm 3}Intel Labs, Cognitive Computing Research \\
    \{xxu,car\}@ir.hit.edu.cn, \{chenfei.wu,nanduan\}@microsoft.com, \{shachar.rosenman,vasudev.lal\}@intel.com
}

\begin{document}

\maketitle

\begin{abstract}
    Vision-Language (VL) models with the \twotower{} architecture have dominated visual-language representation learning in recent years.
    Current VL models either use lightweight uni-modal encoders and learn to extract, align and fuse both modalities simultaneously in a deep cross-modal encoder, or feed the last-layer uni-modal representations from the deep pre-trained uni-modal encoders into the top cross-modal encoder. 
    Both approaches potentially restrict vision-language representation learning and limit model performance.
    In this paper, we propose \methodname{}, which introduces multiple bridge layers that build a connection between the top layers of uni-modal encoders and each layer of the cross-modal encoder. 
    This enables effective bottom-up cross-modal alignment and fusion between visual and textual representations of different semantic levels of pre-trained uni-modal encoders in the cross-modal encoder.
    Pre-trained with only $4$M images, \methodname{} achieves state-of-the-art performance on various downstream vision-language tasks. 
    In particular, on the VQAv2 test-std set, \methodname{} achieves an accuracy of $78.73\%$, outperforming the previous state-of-the-art model \metername{} by $1.09\%$ with the same pre-training data and almost negligible additional parameters and computational costs. 
    Notably, when further scaling the model, \methodname{} achieves an accuracy of $81.15\%$, surpassing models that are pre-trained on orders-of-magnitude larger datasets.
    Code and checkpoints are available at \url{https://github.com/microsoft/BridgeTower}.
\end{abstract}

\section{Introduction}

Vision-Language (VL) tasks aim to perceive, comprehend and fuse both visual and textual information in our complex multi-modal world and then produce cross-modal representations to address difficult cross-modal challenges, such as visual question answering, visual entailment, and image-text retrieval~\citep{balanced_vqa_v2,xie2019visual,young-etal-2014-image}.
Recently, by pre-training on large-scale image-text pairs, cross-modal representations have been improved considerably~\citep{su2019vl,lu2019vilbert,chen2020uniter,zhang2021vinvl,radford2021learning,wang2021simvlm,li2021align,dou2021meter,wang2022OFA,alayrac2022flamingo}. 
Many elaborate Vision-Language Pre-training (VLP) objectives are proposed for mining cross-modal knowledge from image-text pairs, such as Masked Language Modeling (MLM) and Image-Text Matching (ITM).

Most existing VL models can be unified into the \twotower{} architecture, which consists of a visual encoder, a textual encoder, and a cross-modal encoder. 
The models differ in the design of the three encoders.
Benefiting from the rapid progress and prominent performance of Vision Transformer (ViT)~\citep{dosovitskiy2020image} on various vision tasks, recent VL models can adopt ViT as a cross-modal or visual encoder 
without using region features from heavy and time-consuming pre-trained object detectors.

ViLT~\citep{kim2021vilt} adopts linear projection and word embedding as lightweight uni-modal encoders, and uses ViT as the cross-modal encoder to extract, align and fuse the features of both modalities simultaneously.
While parameter-efficient, it may be difficult for ViLT to learn intra- and cross-modal interactions concurrently, and thus its performance lags behind state-of-the-art performance on downstream VL tasks. 
\metername{}~\citep{dou2021meter} uses ViT and RoBERTa~\citep{liu2019roberta} as pre-trained uni-modal encoders and feeds the last-layer uni-modal representations directly into the top cross-modal encoder. 
Although \metername{} achieves performance competitive with the previous region-based state-of-the-art model VinVL~\citep{zhang2021vinvl}, it ignores and wastes different levels of semantic knowledge contained in different layers of pre-trained uni-modal encoders. 
Furthermore, the abstract representations from the last layer of pre-trained uni-modal encoders could be challenging for the top cross-modal encoder to learn cross-modal alignment and fusion~\citep{lu2019vilbert,tan2019lxmert}.

\begin{figure*}[!ht]
    \centering
    \includegraphics[width=0.98\textwidth]{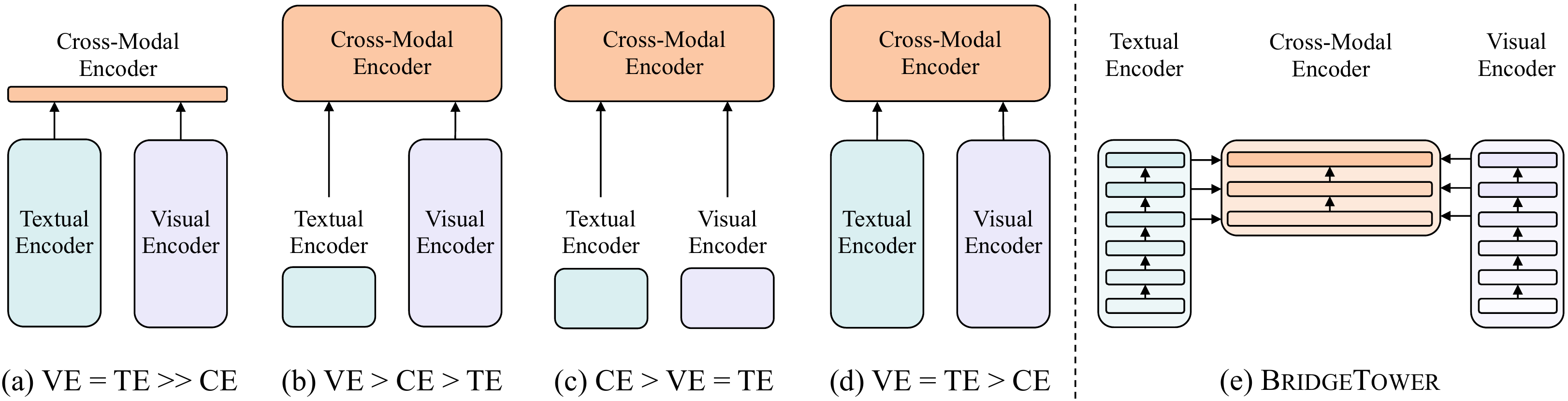}
    \caption{
        (a) -- (d) are four categories of current \twotower{} vision-language models;~(e) gives a brief illustration of the \methodname{} architecture. VE, TE, and CE are short for the Visual Encoder, Textual Encoder, and Cross-modal Encoder, respectively.
        The height of each rectangle represents its relative computational cost. 
        $\text{VE}=\text{TE}$ indicates that the visual encoder and the textual encoder have the same or a similar number of parameters or computational costs. 
        Illustration inspired by ViLT.
    }
    \label{fig:overview}
\end{figure*}

It has been demonstrated that different layers encode different types of information in both vision~\citep{zeiler2014visualizing,dosovitskiy2020image,du2020selective,raghu2021vision,naseer2021intriguing} and language models~\cite{peters-etal-2018-dissecting,liu2019linguistic,jawahar-etal-2019-bert}. 
~\citet{dosovitskiy2020image} and~\citet{raghu2021vision} find that lower layers of ViT attend both locally and globally, while higher layers primarily incorporate global information. 
~\citet{jawahar-etal-2019-bert} find that the intermediate layers of BERT~\citep{devlin-etal-2019-bert} encode a rich hierarchy of linguistic information, starting with surface features at the bottom, syntactic features in the middle, and then semantic features at the top. 
Therefore, it makes perfect sense to utilize multi-layer uni-modal features to obtain effective improvements in both vision~\citep{lin2017feature, huang2017densely, yu2018deep, zheng2021rethinking, xie2021segformer} and language tasks~\citep{peters-etal-2018-deep, wang2018multi, shen2018dense, dou2018exploiting, sun2019fine}. 
The question, therefore, naturally arises: \textit{can we build a bridge between different layers of pre-trained uni-modal encoders and the cross-modal encoder to utilize multi-layer uni-modal features?} 

We propose \methodname{}, a novel transformer-based VL model that takes full advantage of the features of different layers in pre-trained uni-modal encoders.
By introducing multiple bridge layers, the top layers of uni-modal encoders can be connected with each layer of the cross-modal encoder.
This enables effective bottom-up cross-modal alignment and fusion between visual and textual representations of different semantic levels of pre-trained uni-modal encoders in the cross-modal encoder.
Moreover, in principle, the proposed \methodname{} architecture is applicable to any visual, textual or cross-modal encoder.

We conduct extensive experiments on different design choices for \methodname{} and fine-tune it on various downstream VL tasks.
Experimental results show that with only $4$M images for pre-training, our model achieves state-of-the-art performance on various downstream VL tasks, especially $78.73\%$ accuracy on the VQAv2 test-std set, outperforming the previous state-of-the-art model \metername{} by $1.09\%$ with the same pre-training data and almost negligible additional parameters and computational costs.
When further scaling the model, \methodname{} achieves $81.15\%$ accuracy on the VQAv2 test-std set, outperforming models that are pre-trained on orders-of-magnitude larger datasets.

Our contributions are threefold:
\begin{itemize}[noitemsep, nolistsep]
    \item We introduce \methodname{}, a novel transformer-based VL model that achieves substantial improvements over previous state-of-the-art model \metername{} both with and without pre-training.
    \item We propose using multiple bridge layers to connect the top layers of uni-modal encoders with each layer of the cross-modal encoder. Furthermore, we conduct extensive experiments on different design choices for \methodname{}.
    \item We demonstrate the effectiveness of \methodname{} on various VL downstream tasks, including visual question answering (VQAv2), visual entailment (SNLI-VE), and image-text retrieval (Flickr30K) tasks.
\end{itemize}

\section{Related Work}
\subsection{\twotower{} Vision-Language Models}
Following the taxonomy proposed by ViLT~\citep{kim2021vilt}, most VL models can be unified into the \twotower{} architecture shown in Figure~\ref{fig:overview}(a) -- (d). They feed last-layer representations of pre-trained uni-modal encoders into the top cross-modal encoder and can be differentiated by the depth of the textual, visual, and cross-modal encoders\footnote{A cross-modal decoder can be placed on top of the cross-modal encoder, which is omitted since it is not the main study of this paper.}.

CLIP~\citep{radford2021learning} and ALIGN~\citep{jia2021scaling} are representative models that directly perform a shallow fusion (\eg{} dot product) of last-layer representations of equally expressive pre-trained uni-modal encoders in the cross-modal encoder, as illustrated in Figure~\ref{fig:overview}(a).
The remaining models perform deep fusion in the multi-layer transformer-based cross-modal encoder but choose pre-trained uni-modal encoders with varying levels of expressiveness. 
Numerous works~\citep{li2019visualbert,su2019vl,li2020unicoder,chen2020uniter,li2020oscar,zhou2020unified,zhang2021vinvl,pmlr-v139-cho21a,huang2020pixel,huang2021seeing,shen2021much,liu2021kd,li2021unimo,xia2021xgpt,ni2021m3p,chen2022pali,wang2022git,alayrac2022flamingo}
fall in the category of Figure~\ref{fig:overview}(b) as they adopt various types of deep vision models (\eg{} Faster R-CNN~\citep{ren2015faster}, ResNet~\citep{he2016deep} or ViT~\citep{dosovitskiy2020image}) as their visual encoder to obtain region, grid, or patch features, and concatenate them with word embedding to feed into their top cross-modal encoder.
The third category of models~\citep{kim2021vilt, wang2021simvlm, wang2021distilled, wang2022OFA}, illustrated in Figure~\ref{fig:overview}(c), utilizes lightweight visual and lightweight textual encoders and handles both modalities in a single transformer-based cross-modal encoder. 
In contrast, models~\citep{lu2019vilbert, tan2019lxmert, kamath2021mdetr, li2021align, zeng2021multi, dou2021meter, wang2021vlmo, li2022blip, li2022unimo, wang2022image, yu2022coca, li2022mplug}, which belong to Figure~\ref{fig:overview}(d) category, use expressive deep pre-trained uni-modal encoders and feed their last-layer representation into the top multi-layer cross-modal encoder.

Regardless of the visual, textual, or cross-modal encoders they utilize, most current models ignore the various levels of semantic information at the different layers of pre-trained uni-modal encoders, and simply utilize the last-layer uni-modal representations for cross-modal alignment and fusion. 
While the models belonging to Figure~\ref{fig:overview}(c) appear to retain the possibility of utilizing different levels of uni-modal semantic information, it could be challenging for them to learn intra- and cross-modal interactions concurrently without modality-specific parameters. Their unconstrained cross-modal interaction could impede intra-modal interaction~\citep{dou2021meter, du2022survey}.
This may be the reason why the performance of ViLT lags behind models in the Figure~\ref{fig:overview}(d) category, and why SimVLM~\citep{wang2021simvlm} and OFA~\citep{wang2022OFA} need to use significantly more data to obtain competitive performance compared with \metername{}.

Unlike current models, \methodname{}, as shown in Figure~\ref{fig:overview}(e), proposes using multiple bridge layers to connect the top layers of uni-modal encoders with each layer of the cross-modal encoder.
This does not affect intra-modal interaction in the pre-trained uni-modal encoders, and enables different semantic levels of visual and textual representations to interact thoroughly and mildly in the bottom-up directions at each layer of the cross-modal encoder.

\subsection{Multi-Layer Feature Utilization}
Multi-layer feature utilization has been demonstrated to be an effective method of making full use of the information contained in different layers of neural networks to improve the representation and generalization capabilities of computer vision~\citep{ronneberger2015u, liu2016ssd, lin2017feature, huang2017densely, yu2018deep, kirillov2019panoptic, zheng2021rethinking, xie2021segformer, naseer2021intriguing}, natural language processing~\citep{peters-etal-2018-deep, wang2018multi, shen2018dense, dou2018exploiting, jawahar-etal-2019-bert, sun2019fine, dou2019dynamic} and multi-modal models~\citep{dou2021meter,nagrani2021attention}.

Since \citet{zeiler2014visualizing} introduce a visualization technique and find that different patterns are learned in different layers of CNN models, then many researchers exploit features of different layers to improve detection and semantic segmentation. 
U-Net~\citep{ronneberger2015u} and FPN~\citep{lin2017feature} propose to adopt lateral connections for associating feature maps from different layers across resolutions and semantic levels. 
The same idea is also applicable to ViT-based models. SETR~\citep{zheng2021rethinking} and SegFormer~\citep{xie2021segformer} aggregate features from different layers to improve semantic segmentation performance.
In natural language processing, researchers~\citep{sogaard-goldberg-2016-deep, hashimoto2017joint, belinkov2017neural, peters-etal-2018-dissecting, jawahar-etal-2019-bert, liu2019linguistic} find that Recurrent Neural Networks (RNN)~\citep{hochreiter1997long} and BERT~\citep{devlin-etal-2019-bert} encode different types of semantic information in different layers. Hence, ~\citet{peters-etal-2018-deep} and~\citet{sun2019fine} use the concatenation or weighted sum of representations from different layers of RNN or BERT as input for different task heads. 
~\citet{dou2018exploiting} explore layer aggregation with multi-layer attention mechanisms.

Recent multi-modal models exploit features from different layers. 
MBT~\citep{nagrani2021attention} introduces simple bottleneck tokens at multiple layers to jointly model intra- and restricted cross-modal correlations.
While MBT achieves good performance on audio-visual benchmarks, learning complex vision-language alignment and fusion via a limited number of bottleneck tokens instead of a cross-modal encoder maybe too difficult, which limits cross-modal alignment.
\metername{} feeds the weighted sum of representations from each layer of the bottom uni-modal encoder into the top cross-modal encoder; they find this can improve performance by a small margin without VLP but can degrade performance with VLP.

In a departure from existing models, \methodname{} considers detailed interactions between the top layers of uni-modal encoders and each layer of the cross-modal encoder.
It is intuitive to connect pre-trained uni-modal encoders and the cross-modal encoder via multiple bridge layers, in order to achieve comprehensive cross-modal alignment and fusion of the uni-modal representations of different semantic levels.
Most importantly, unlike the simple multi-layer feature fusion method in \metername{}, \methodname{} can significantly improve performance both with and without vision-language pre-training on large-scale image-text data.

\section{Approach}
\begin{figure*}[t]
  \centering
  \includegraphics[width=0.85\textwidth]{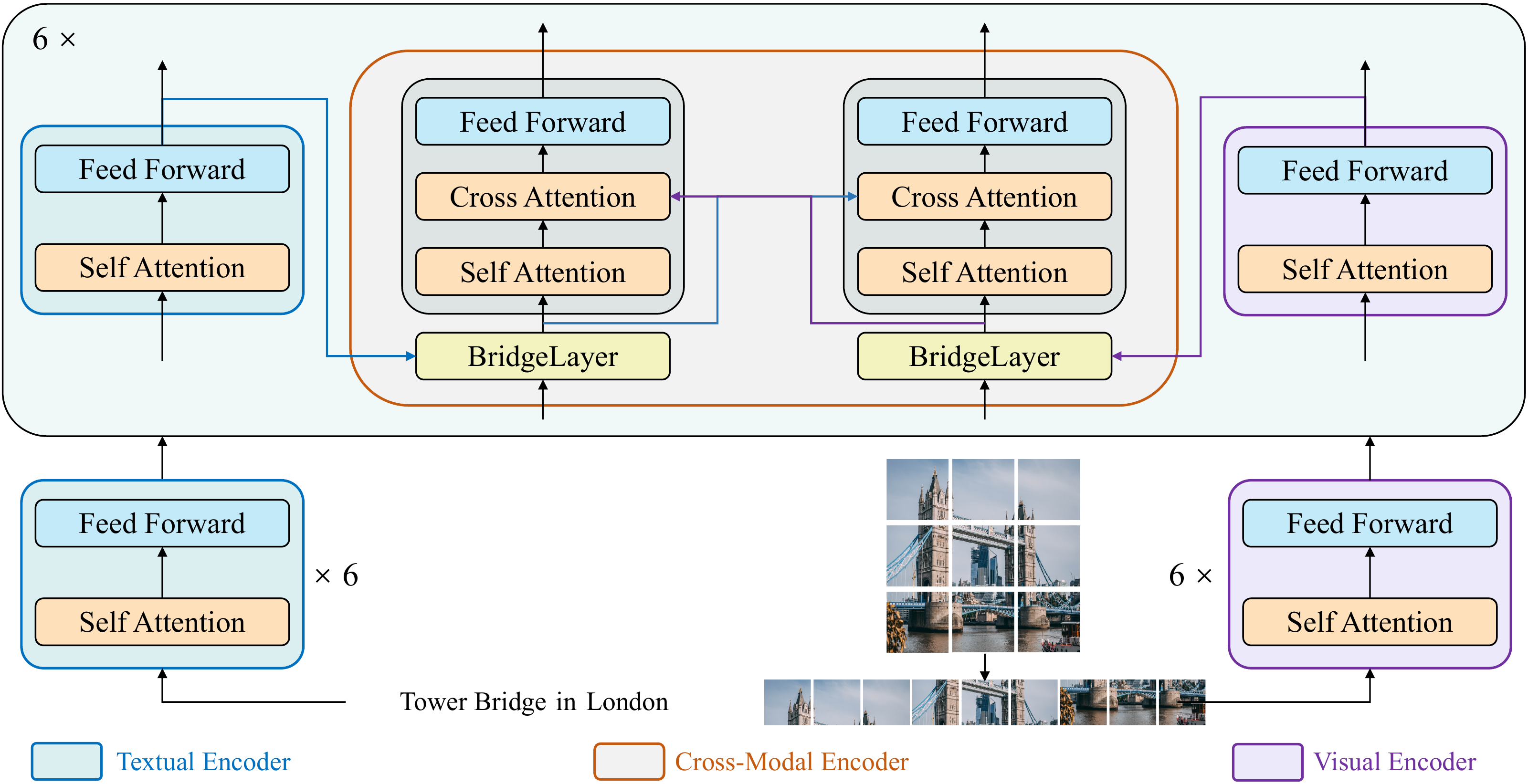}
  \caption{
      Illustration of \methodname{}.
      \methodname{} consists of a $12$-layer textual encoder, a $12$-layer visual encoder, and a 6-layer cross-modal encoder.
      Each of the top $6$ layers of the visual and textual encoders is connected to each layer of the cross-modal encoder via {bridge layers}, which brings {bottom-up} alignment and fusion to the cross-modal encoder.
  }
  \label{fig:framework}
\end{figure*}

As shown in Figure~\ref{fig:framework}, 
\methodname{} consists of a visual encoder, a textual encoder and a cross-modal encoder with multiple lightweight bridge layers. 
Our goal is to build a bridge between each uni-modal encoder and the cross-modal encoder to enable comprehensive and detailed interaction at each layer of the cross-modal encoder. 
Our goal is not to develop new encoders; in principle, one can apply any visual, textual, or cross-modal encoder in the proposed architecture.

\subsection{Visual Encoder}
Recent works~\citep{shen2021much, dou2021meter} show that CLIP's visual encoder has strong capabilities on VL tasks.
We follow METER to adopt CLIP-ViT-B/16 as the pre-trained visual encoder.
For each input 2D image $\mathbf{I} \in \mathbb{R}^{H \times W \times C}$, where $(H, W)$ is the resolution of the input image and $C$ is the number of channels, ViT reshape it into a sequence of flattened 2D patches $\mathbf{P} \in \mathbb{R}^{N \times (P^2C)}$, where $(P, P)$ is the image patch resolution and $N=\frac{HW}{P^2}$ is the number of patches. 
Similar to BERT, ViT also prepends the \texttt{[class]} token to the patch sequence and uses learnable 1D position embeddings $\mathbf{V}^{pos} \in \mathbb{R}^{(N+1) \times D_v}$, where $D_v$ is the dimension of the visual encoder. 
The input visual representation can be calculated as follows:
\begin{equation}
    \mathbf{V}_0=[\mathbf{E}_{\texttt{[class]}};\mathbf{p}_1 \mathbf{W_p};\dots;\mathbf{p}_N \mathbf{W_p}] + \mathbf{V}^{pos},
\end{equation}
where $\mathbf{W_p} \in \mathbb{R}^{(P^2C) \times D_v}$ is the trainable linear projection layer and $\mathbf{V}_0 \in \mathbb{R}^{(N+1) \times D_v}$.
Each layer of ViT consists of a multi-head self-attention (MSA) block and a feed-forward network (FFN) block. We omit the computation details and simplify them as $\operatorname{Encoder}^\mathrm{V}$. The $\ell$-th layer representation can be denoted as $\mathbf{V}_{\ell} = \operatorname{Encoder}^\mathrm{V}_\ell(\mathbf{V}_{\ell-1}), \ell = 1 \dots L_V$, where $L_V$ is the number of layers of the visual encoder.

\subsection{Textual Encoder}
Since RoBERTa achieves robust performance on a wide range of NLP tasks, we adopt RoBERTa$_{\text{BASE}}$ as our textual encoder. Each input sequence $\mathbf{w}$ is tokenized by the byte-level Byte-Pair Encoding (BPE)~\citep{sennrich-etal-2016-neural, radford2019language}. $\texttt{[<s>]}$ token and $\texttt{[</s>]}$ token are added to the sequence as the start and end tokens, respectively.
The input textual representation can be represented as:
\begin{equation}
    \mathbf{T}_0=[\mathbf{E}_{\texttt{[<s>]}};\mathbf{E}_{{w}_1};\dots;\mathbf{E}_{{w}_M};\mathbf{E}_{\texttt{[</s>]}}] + \mathbf{T}^{pos},
\end{equation}
where $\mathbf{T}_{0} \in \mathbb{R}^{(M+2) \times D_t},\mathbf{E}$ is the word embedding matrix, $M$ is the number of tokens, $D_t$ is the dimension of the textual encoder, and $\mathbf{T}^{pos}$ is the positional embedding matrix. 
Similarly, we denote the $\ell$-th layer of the textual encoder as $\operatorname{Encoder}^\mathrm{T}_\ell$, and the $\ell$-th layer representation can be denoted as $\mathbf{T}_{\ell} = \operatorname{Encoder}^\mathrm{T}_\ell(\mathbf{T}_{\ell-1}), \ell = 1 \dots L_T$, where $L_T$ is the number of layers of the textual encoder.

\subsection{Cross-Modal Encoder with Bridge Layers} 
\label{sec:bridge-layers}
\citet{hendricks2021decoupling} perform analysis on different types of attention mechanisms used in the existing transformer-based cross-modal encoders and demonstrate that the \coattention{} mechanism~\citep{lu2019vilbert} performs best. 
This mechanism uses a different set of parameters for each modality. 
For example, for the visual part of the cross-modal encoder, the queries of each MSA block are from the visual modality. However, the keys and values are from the other modality (\ie{} the textual modality). 
We, therefore, adopt the \coattention{} mechanism. 
Formally, we define the $\ell$-th layer of the cross-modal encoder as $\operatorname{Encoder}^\mathrm{Z}_\ell$, which consists of a visual part and a textual part. 
Each part consists of an MSA block, a multi-head cross-attention (MCA) block, and an FFN block. 
For brevity, the interactions at each layer are defined as:
\begin{gather}
    \tilde{\mathbf{Z}}^\mathrm{V}_{\ell} = {\mathbf{Z}}^\mathrm{V}_{\ell-1}, \label{eq:prev1} \\
    \tilde{\mathbf{Z}}^\mathrm{T}_{\ell} = {\mathbf{Z}}^\mathrm{T}_{\ell-1}, \label{eq:prev2} \\
    \mathbf{Z}^\mathrm{V}_{\ell}, \mathbf{Z}^\mathrm{T}_{\ell}= \operatorname{Encoder}^\mathrm{Z}_\ell(\tilde{\mathbf{Z}}^\mathrm{V}_{\ell}, \tilde{\mathbf{Z}}^\mathrm{T}_{\ell}), \ell = 1 \dots L_Z,
\end{gather}
where $\mathbf{Z}^\mathrm{\{\mathrm{V},\mathrm{T}\}}_{\ell}$ is the output representation of the visual or textual part at the $\ell$-th layer, $\tilde{\mathbf{Z}}^\mathrm{\{\mathrm{V},\mathrm{T}\}}_{\ell}$ is the input of each part, and $L_Z$ is the number of layers of the cross-modal encoder.

Generally, current VL models, such as \metername{}, directly use the output representation of the previous layer as the input to $\operatorname{Encoder}^\mathrm{Z}_\ell$ (Equation~\ref{eq:prev1}~\&~\ref{eq:prev2}). 
$\mathbf{Z}^\mathrm{V}_{0}, \mathbf{Z}^\mathrm{T}_{0}$ are initialized with the last-layer representations from pre-trained uni-modal encoders:  $\mathbf{Z}^\mathrm{V}_{0} = \mathbf{V}_{L_V}\mathbf{W}_V + \mathbf{V}^{type}, \mathbf{Z}^\mathrm{T}_{0} = \mathbf{T}_{L_T}\mathbf{W}_T + \mathbf{T}^{type}$, where $\mathbf{W}_V \in \mathbb{R}^{D_V \times D_Z}$ and $\mathbf{W}_T \in \mathbb{R}^{D_T \times D_Z}$ are linear projections, $\mathbf{V}^{type}$ and $\mathbf{T}^{type}$ are the modality type embeddings.

However, in this paper, we propose using multiple bridge layers to connect the top layers of uni-modal encoders with each layer of the cross-modal encoder: 
\begin{gather}
    \tilde{\mathbf{Z}}^\mathrm{V}_{\ell} = \operatorname{BridgeLayer}^\mathrm{V}_\ell(\mathbf{Z}^\mathrm{V}_{\ell-1}, \mathbf{V}_{k}\mathbf{W}_V + \mathbf{V}^{type}), \\
    \tilde{\mathbf{Z}}^\mathrm{T}_{\ell} = \operatorname{BridgeLayer}^\mathrm{T}_\ell(\mathbf{Z}^\mathrm{T}_{\ell-1}, \mathbf{T}_{k}\mathbf{W}_T + \mathbf{T}^{type}),
\end{gather}
where $k$ denotes the index of layer representations of uni-modal encoders. 
In this paper, $L_V=L_T=12, L_Z=6$ and we use the representations of the top $6$ layers of uni-modal encoders, which means that $k=7,\dots,12$. Take the input of $\operatorname{Encoder}^\mathrm{Z}_2$ as an example:
\begin{gather}
    \tilde{\mathbf{Z}}^\mathrm{V}_{2} = \operatorname{BridgeLayer}^\mathrm{V}_{2}(\mathbf{Z}^\mathrm{V}_{1}, \mathbf{V}_{8}\mathbf{W}_V + \mathbf{V}^{type}), \\
    \tilde{\mathbf{Z}}^\mathrm{T}_{2} = \operatorname{BridgeLayer}^\mathrm{T}_{2}(\mathbf{Z}^\mathrm{T}_{1}, \mathbf{T}_{8}\mathbf{W}_T + \mathbf{T}^{type}).
\end{gather}

Utilizing our proposed bridge layers, top-layer uni-modal representations can be bridged with each layer of the cross-modal encoder, thus incorporating different semantic levels of uni-modal representations into cross-modal interaction.
In the spirit of ResNet and Transformer~\citep{vaswani2017attention}, the simplest formal definition of a bridge layer is:
\begin{equation}
    \operatorname{BridgeLayer}(x, y) = \operatorname{LayerNorm}(x + y).
\end{equation}

In Sec.~\ref{sec:investigation_and_analysis}, we describe the extensive experiments we conducted on different design choices for \methodname{}, including the formal definition of bridge layers and the number of cross-modal layers.

\subsection{Pre-training Objectives}
We pre-train \methodname{} with two commonly used vision-language pre-training objectives: MLM and ITM.

\paragraph{Masked Language Modeling.} 
MLM is a common objective for language and vision-language pre-training. Given an image-text pair, following UNITER~\citep{chen2020uniter}, we use conditional masking for MLM, which means we randomly mask $15\%$ of tokens in the token sequence while keeping the input image patch sequence untainted. The model is trained to reconstruct the original tokens conditioned on incomplete input token sequence and complete observed image patch sequence. We adopt the same masking strategy and MLM task head as RoBERTa. The last-layer representation of the textual part of the cross-modal encoder is used as input to the MLM task head.

\paragraph{Image-Text Matching.}
ITM aims to predict whether the given image-text pair is positive (matched) or negative (mismatched). Matched and mismatched image-text pairs are fed into our model with the same probability. We pass the final representations of $\texttt{[class]}$ and $\texttt{[<s>]}$ token in the cross-modal encoder to the non-linear layer activated by $\texttt{Tanh}$, respectively. The concatenation of the outputs is fed into a linear classifier with cross-entropy loss for binary classification. \looseness=-1

\subsection{Fine-Tuning on Downstream Tasks}
For visual question answering and visual entailment, we use the same strategy as for ITM.
For image-text retrieval, following ALBEF~\citep{li2021align}, our model is jointly optimized with image-text contrastive (ITC) loss and ITM loss.
Two linear projections are added on top of both uni-modal encoders to obtain uni-modal representations of image-text pairs, and then compute their contrastive similarity by dot product. 
Then, instead of randomly sampling negatives for the ITM task, for each image (text) in a mini-batch, we use the contrastive similarity distribution from the ITC task to sample one hard in-batch negative text (image).
In inference, we first compute the contrastive similarity for all images and texts, and then take the top-k candidates and calculate their ITM scores for ranking.

\section{Experiment}

\subsection{Implementation Details}
\label{sec:implementation_details}
\methodname{} consists of a pre-trained textual encoder, RoBERTa\modelbase{} with $124$M parameters, a pre-trained visual encoder, CLIP-ViT-B-224/16 with $86$M parameters, and a random-initialized $6$-layer cross-modal encoder with $113$M parameters. For each layer of the cross-modal encoder, the hidden size is set to $768$, the intermediate size of feed-forward networks is set to $3,072$, and the number of heads is set to $12$. The maximum length of the text sequence is set to $50$.
The patch size is $16 \times 16$. Center-crop is used to resize each input image to the same resolution, and we also apply RandAugment~\citep{cubuk2020randaugment} to the input images following previous works~\citep{li2021align,dou2021meter}. 
We use the AdamW~\citep{loshchilov2018decoupled} optimizer with a base learning rate of $2e^{-5}$ and weight decay of $0.01$. The learning rate is warmed up for $10\%$ of the total training steps and then decayed linearly to zero for the rest of the training steps.
Following \metername{}, the learning rate of the cross-modal encoder is five times higher than that of uni-modal encoders.

We evaluate \methodname{} by fine-tuning the entire model on the visual question answering (VQAv2)~\citep{balanced_vqa_v2}, visual entailment (SNLI-VE)~\citep{xie2019visual}, and image-text retrieval (Flickr30K)~\citep{young-etal-2014-image} tasks.
We use an image resolution of $384 \times 384$ for these downstream VL tasks, except for VQAv2, where we use $576 \times 576$ for a robust evaluation and fair comparison with \metername{}. 
Standard settings and splits are used for all datasets. For VQAv2, where we follow the common practice~\citep{balanced_vqa_v2, teney2018tips}: convert VQAv2 to a classification task with $3,129$ answer classes; train the model with training data and validation data, and evaluate the model on the test-dev data.\looseness=-1

\subsection{Investigation and Analysis}
\label{sec:investigation_and_analysis}
In this section, we evaluate different design choices for \methodname{} on the VQAv2 and Flickr30K datasets. 
Each model is initialized with CLIP-ViT-B-224/16 and RoBERTa\modelbase{} pre-trained weights, and then directly fine-tuned on the two downstream tasks without VLP.
All experimental settings are the same as \metername{} for fair comparisons.
In our preliminary experiments, the uni-modal representations of the top layers perform much better than the middle and bottom layers.
Thus, we use the top $6$ layer representations of the uni-modal encoders as the corresponding inputs for each bridge layer in the bottom-up directions.

\subsubsection{Design Choice \uppercase\expandafter{\romannumeral1}: Formal Definition of Bridge Layers}

Table~\ref{tab:design-choice-1} shows, perhaps unexpectedly but not very surprisingly, 
that row~(a) provides the best results using the minimum number of parameters and achieves an accuracy of $75.18\%$ on the test-dev set of VQAv2 and RSUM of $533.84$ on the test set of Flickr30K. 
The additional parameters used for interpolation cause slight performance degradation in rows~(c)\&(d).
Rows~(e) -- (h) try to incorporate generally used feature transformation forms into the bridge layer, but the additional computation and parameters instead lead to performance degradation.
Inspired by ResNet and ViTDet~\citep{li2022exploring}, in row~(i), we incorporate row~(a) into row~(e) as the residual connection. $\mathbf{W}_*$ is initialized as zero so that row~(i) is initially equivalent to row~(a). 
Although the performance is significantly higher than row~(e) ($74.55 \rightarrow 75.10$), there is no significant gain compared with row~(a). 
Hence, we choose row~(a) (\texttt{Add\&Norm}) as the default bridge layer.

\begin{table}[t]
    \tablestyle{1.5pt}{1.0}
    \scalebox{1.0}{
        \begin{tabular}{lc|c|c}
            $\operatorname{BridgeLayer}(x, y)$ & \#~Params & {Test-Dev} & RSUM \\
            \shline
            (a) $x + y$ & \textbf{18.4K} & \textbf{75.18} & \textbf{533.8} \\
            (b) $x \odot y$ & 18.4K & 73.41 & 530.4 \\
            (c) $\alpha x + \left(1-\alpha\right) y, \alpha \in \mathbb{R}^{D_Z}$ & 26.0K & 75.09 & 532.9 \\
            (d) $\alpha x + \left(1-\alpha\right) y, \alpha = \sigma{\left(\mathbf{W}\left[x;y\right]\right)}$ & 11.8M & 75.13 & 533.1 \\
            (e) $\mathbf{W}\left[x;y\right]$ & 11.8M & 74.55 & 532.2 \\
            (f) $\mathbf{W}_2\left(\operatorname{GeLU}\left(\mathbf{W}_1[x;y]\right)\right)$ & 35.4M & 74.26 & 530.2 \\
            (g) $\operatorname{MCA}\left(x ,y\right)$ & 23.6M & 73.67 & 514.3 \\
            (h) $\operatorname{FFN}\left(\operatorname{MCA}\left(x, y\right)\right)$ & 70.8M & 73.54 & 511.1 \\
            (i) $x + y + \mathbf{W}_*\left[x;y\right]$ & 11.8M & 75.10 & 533.1\\ 
        \end{tabular}
    }
    \caption{
        Performance and number of parameters for different formal definitions of bridge layers. We omit the layer normalization used in each form. 
        $x$ denotes the output cross-modal representation of the previous layer and $y$ denotes the corresponding input uni-modal representation. 
        RSUM indicates the sum of recall metrics for image-text retrieval.
    }
    \label{tab:design-choice-1}
\end{table}

\subsubsection{Design Choice \uppercase\expandafter{\romannumeral2}: Number of Cross-Modal Layers}
\label{sec:design-choice-2}

In \methodname{}, the cross-modal encoder is not located on top of the uni-modal encoders, but between them. Each cross-modal layer is connected to the corresponding layer of the uni-modal encoder by a bridge layer. 
Therefore, based on the two $12$-layer uni-modal encoders we used, the number of cross-modal layers can be $[1, 12]$.
Table~\ref{tab:design-choice-2} shows the performance of \methodname{} with different numbers of cross-modal layers.
It is illuminating to note that adding more cross-modal layers does not constantly improve performance, possibly because
($i$) more cross-modal layers are more difficult to train and are more data-hungry;
($ii$) uni-modal representations of top layers are beneficial to cross-modal alignment and fusion, while uni-modal representations of bottom layers may be less useful and even detrimental.
We also evaluate \metername{} and find that while the only difference between the two models is the bridge layers, \methodname{} can achieve consistent performance gains for different numbers of cross-modal layers.
It further illustrates that the bridge layers can facilitate effective cross-modal alignment and fusion with uni-modal representations of different semantic levels in the cross-modal encoder.

\subsubsection{Apply Different Visual and Textual Backbones}
We apply different visual and textual backbones as pre-trained uni-modal encoders and directly fine-tune on downstream tasks to further investigate the impact brought by bridge layers. 
As shown in Table~\ref{tab:different-encoders}, no matter what visual and textual encoders we apply, the performances of \methodname{} are consistently and significantly better than that of \metername{}. 
This further demonstrates the effectiveness of our proposed \methodname{} architecture and bridge layers for vision-language representation learning.

\subsection{Comparison with Previous Arts}
In this section, we describe how to pre-train \methodname{} with the best-performing setting (Sec.~\ref{sec:investigation_and_analysis})
and compare its fine-tuning performance with previous works.

\paragraph{Pre-training Setup.}
\label{sec:implementation_details_2}
We use four public image-caption datasets for pre-training: Conceptual Captions (CC)~\citep{sharma-etal-2018-conceptual}, SBU Captions~\citep{NIPS2011_5dd9db5e}, MSCOCO Captions~\citep{chen2015microsoft}, and Visual Genome (VG)~\citep{krishna2017visual}. The total number of unique images in the combined data is $4$M.
We pre-train \methodname{} for $100$k steps on $8$ NVIDIA A100 GPUs with a batch size of $4,096$.
All the pre-training settings for \methodname{} are the same as for \metername{} for a fair comparison.
The learning rate is set to $1e^{-5}$. No data augmentation is used except for center-crop~\citep{radford2021learning,dou2021meter}. The image resolution in pre-training is set to $288 \times 288$. Other hyperparameters remain unchanged based on the experiments in Sec.~\ref{sec:investigation_and_analysis}.

\begin{table}[t]
    \tablestyle{3.3pt}{1.0}
    \scalebox{1}{
        \begin{tabular}{rr|cl|cl}
            \multirow{2}{*}{$L_Z$} & \multirow{2}{*}{\#~Params} & \multicolumn{2}{c|}{VQAv2 Test-Dev} & \multicolumn{2}{c}{Flickr30K RSUM} \\
            &  & \metername{} & \multicolumn{1}{c|}{Ours} & \metername{} & \multicolumn{1}{c}{Ours} \\
            \shline
            2 & 37.8M & 72.84 &  74.12 ($\uparrow$ 1.28) & 526.0 &  527.1 ($\uparrow$ 1.1) \\
            3 & 56.8M & 73.47 &  74.36 ($\uparrow$ 0.89) & 526.5 &  528.6 ($\uparrow$ 2.1) \\
            4 & 75.6M & 73.71 &  75.00 ($\uparrow$ 1.29) & 527.9 &  529.7 ($\uparrow$ 1.8) \\
            5 & 94.6M & 73.80 &  74.98 ($\uparrow$ 1.18) & 528.8 &  531.8 ($\uparrow$ 3.0) \\
            6 & 113.4M & 74.04 &  \textbf{75.18} ($\uparrow$ 1.14) & 530.7 &  \textbf{533.8} ($\uparrow$ 3.1) \\
            8 & 151.2M & 73.97 &  75.07 ($\uparrow$ 1.10) & 530.0 &  531.6 ($\uparrow$ 1.6) \\
            10 & 189.0M & 73.45 & 75.06 ($\uparrow$ 1.61) & 529.6 & 531.7 ($\uparrow$ 2.1) \\
            12 & 226.8M & 71.88 & 74.94 ($\uparrow$ 3.06) & 528.7 & 531.4 ($\uparrow$ 2.7)\\
        \end{tabular}
    }
    \caption{
        Performance of \metername{} and \methodname{} with different number of cross-modal layers. \#~Params denotes the number of parameters of the cross-modal encoder.
    }
    \label{tab:design-choice-2}
\end{table}

\begin{table}[t]
    \tablestyle{2pt}{1.0}
    \scalebox{0.88}{
        \begin{tabular}{cc|cc|cc}
            {Visual} & {Textual} & \multicolumn{2}{c|}{VQAv2 Test-Dev} & \multicolumn{2}{c}{Flickr30K RSUM} \\
            Backbone & Backbone & \metername{} & \multicolumn{1}{c|}{Ours} & \metername{} & \multicolumn{1}{c}{Ours} \\
            \shline
        DeiT B-224/16 & RoBERTa & 69.98 & 70.83 ($\uparrow$ 0.85) & 448.0 & 455.7 ($\uparrow$ 7.7) \\
        ViT B-224/16 & RoBERTa & 70.26 & 72.24 ($\uparrow$ 1.98) & 472.7 & 476.9 ($\uparrow$ 4.2) \\
        ViT B-384/16 & RoBERTa & 70.52 & 72.38 ($\uparrow$ 1.86) & 472.8 & 477.1 ($\uparrow$ 4.3) \\
        CLIP-VIT-B/32 & RoBERTa & 72.19 & 72.91 ($\uparrow$ 0.72) & 508.8 & 512.0 ($\uparrow$ 3.2) \\
        CLIP-VIT-B/16 & BERT & 74.09 & 74.89 ($\uparrow$ 0.80) & 522.1 & 526.5 ($\uparrow$ 4.4) \\
        CLIP-VIT-B/16 & RoBERTa & 74.04 & \textbf{75.18} ($\uparrow$ 1.14) & 530.7 & \textbf{533.8} ($\uparrow$ 3.1) \\
        \end{tabular}
    }
    \caption{
        Performance of \metername{} and \methodname{} with different visual and textual encoders. The image resolution of all CLIP visual backbones is $224 \times 224$.
    }
    \label{tab:different-encoders}
\end{table}

\begin{table*}[ht]
    \tablestyle{5pt}{1.0} 
    \scalebox{1.0}{
      \begin{tabular}{lrc|ccccc}
        \multirow{2}{*}{Model} & {\#~Pre-train} & {Visual} & {Test-Dev} & \multicolumn{4}{c}{Test-Standard}  \\
          & Images~ & Backbone & Overall & Yes/No & Number & Other & Overall \\
        \shline
      \multicolumn{7}{l}{ { \it{Base-Size Models} } }\\
      \hline
       ViLT\modelbase{}~\citep{kim2021vilt} & 4M & ViT-B-384/32 & 71.26 & - & - & - & - \\
      UNITER\modelbase{}~\citep{chen2020uniter}\,$\ast$ & 4M & Faster R-CNN & 72.70 & - & - & - & 72.91   \\
      VILLA\modelbase{}~\citep{gan2020large}\,$\ast$ & 4M & Faster R-CNN & 73.59 & - & - & - & 73.67 \\
      UNIMO\modelbase{}~\citep{li2021unimo} & 4M & Faster R-CNN & 73.79 & - & - & - & 74.02  \\
      ALBEF\modelbase{}~\citep{li2021align}\,$\ast$ & 4M & DeiT-B-224/16 & {74.54} & - & - & - & {74.70} \\
      ALBEF\modelbase{}~\citep{li2021align}\,$\ast$ & 14M & DeiT-B-224/16 & {75.84} & - & - & - & {76.04} \\
      VinVL\modelbase{}~\citep{zhang2021vinvl} & 5.7M & ResNeXt-152 & {75.95} & - & - & - & 76.12 \\
      \vlmoname{}\modelbase{}~\citep{wang2021vlmo} & 4M & BEiT-B-224/16 & 76.64 & - & - & - & 76.89 \\
      BLIP\modelbase{}~\citep{li2022blip}\,$\ast$ & 14M & DeiT-B-224/16 & 77.54 & - & - & - & 77.62 \\
      \metername{}\modelbase{}~\citep{dou2021meter} & 4M & CLIP-ViT-B-224/16 & 77.68 & 92.49 & 58.07 & 69.20 & 77.64 \\
      mPLUG~\citep{li2022mplug}\,$\ast$ & 4M & CLIP-ViT-B-224/16 & 77.94 & - & - & - & 77.96 \\
      OFA\modelbase{}~\citep{wang2022OFA}\,$\ast$\,$\star$ & 54M & ResNet-101 & 77.98 & - & - & - & 78.07  \\
      SimVLM\modelbase{}~\citep{wang2021simvlm}\,$\star$ & 1.8B & ResNet-101 & 77.87 & - & - & - & 78.14 \\
      BLIP\modelbase{}~\citep{li2022blip}\,$\ast$ & 129M & DeiT-B-224/16 & 78.24 & - & - & - & 78.17 \\
      \methodname{}\modelbase{} (\textbf{Ours})  & \textbf{4M} & CLIP-ViT-B-224/16 & \textbf{78.66} & \textbf{92.92} & \textbf{60.69} & \textbf{70.51} & \textbf{78.73} \\
      \methodname{}\modelbase{} (\textbf{Ours})\,$\ast$  & \textbf{4M} & CLIP-ViT-B-224/16 & \textbf{79.10} & \textbf{93.06} & \textbf{62.19} & \textbf{70.69} & \textbf{79.04} \\
      \hline
      \multicolumn{7}{l}{ { \it{Large-Size Models} } }\\
      \hline
      UNITER\modellarge{}~\citep{chen2020uniter}\,$\ast$ & 4M & Faster R-CNN & 73.82 & - & - & - & 74.02   \\
      VILLA\modellarge{}~\citep{gan2020large}\,$\ast$ & 4M & Faster R-CNN & 74.69 & - & - & - & 74.87 \\
      UNIMO\modellarge{}~\citep{li2021unimo} & 4M & Faster R-CNN & 75.06 & - & - & - & 75.27  \\
      VinVL\modellarge{}~\citep{zhang2021vinvl} & 5.7M & ResNeXt-152 & {76.52} & 92.04 & 61.50 & 66.68 & 76.63 \\
      {SimVLM\modellarge{}~\citep{wang2021simvlm}} & 1.8B & ResNet-152 & 79.32 & - & - & - & 79.56 \\
      \vlmoname{}\modellarge{}~\citep{wang2021vlmo} & 4M & BEiT-L-224/16 & 79.94 & - & - & - & 79.98 \\
      OFA\modellarge{}~\citep{wang2022OFA}\,$\ast$\,$\star$ & 54M & ResNet-152 & 80.43 & 93.32 & \textbf{67.31} & 72.71 & 80.67 \\
      \methodname{}\modellarge{} (\textbf{Ours})  & \textbf{4M} & CLIP-ViT-L-224/14 & \textbf{81.25}  & \textbf{94.69} & {64.58} & \textbf{73.16} & \textbf{81.15} \\
      \methodname{}\modellarge{} (\textbf{Ours})\,$\ast$  & \textbf{4M} & CLIP-ViT-L-224/14 & \textbf{81.52}  & \textbf{94.80} & {66.01} & \textbf{73.45} & \textbf{81.49} \\
      \hline
    \multicolumn{7}{l}{\it{Huge or even Larger Size Models}}\\
    \hline
      {SimVLM\modelhuge{}~\citep{wang2021simvlm}} & 1.8B & Larger ResNet-152 & {80.03} & 93.29 & 66.54 & 72.23 & {80.34} \\
      \metername{}\modelhuge{}~\citep{dou2021meter} & 14M & Florence-CoSwin-H & 80.33 & 94.25 & 64.37 & 72.30 & 80.54 \\
      mPLUG~\citep{li2022mplug}\,$\ast$ & 14M & CLIP-ViT-L-224/14 & 81.27 & - & - & - & 81.26 \\
      GIT2~\citep{wang2022git}\,$\ast$\ & 10.5B & DaViT(4.8B) & 81.74 & 92.90 & 67.06 & 75.77 & 81.92 \\
      OFA\modelhuge{}~\citep{wang2022OFA}\,$\ast$\,$\star$ & 54M & ResNet-152 & 82.0 & 94.66 & 71.44 & 73.35 & 81.98 \\
      Flamingo~\citep{alayrac2022flamingo}\,$\star$ & 2.3B & NFNet-F6 & 82.0 & - & - & - & 82.1 \\
      CoCa~\citep{yu2022coca}\,$\star$ & 4.8B & ViT-G-288/18 & 82.3 & 94.55 & 70.25 & 74.46 & 82.33 \\
      BEiT-3~\citep{wang2022image} & 28M & BEiT-3 & 84.19 & \textbf{96.43} & \textbf{73.63} & 75.92 & 84.18 \\
      PaLI~\citep{chen2022pali} & 1.6B & ViT-E-224 & \textbf{84.3} & 96.13 & 69.07 & \textbf{77.58} & \textbf{84.34} \\
    \end{tabular}
    }
    \caption{
        Comparisons with previous models on visual question answering (VQAv2). The best score is bolded. The models are divided into base size and large/huge size. B, N and M in ViT-B-N/M denote the model size, image resolution and patch size, respectively. $\ast$ indicates that the model also uses VG-QA data to fine-tune on VQAv2. $\star$ denotes the model is trained from scratch. ``\# Pre-train Images'' denotes the number of images in VLP (the images for pre-trained visual and textual backbones are not counted). \looseness=-1
    }
    \label{tab:vqav2_results}
\end{table*}

\begin{table*}[!h]
    \tablestyle{5pt}{0.99}
    \scalebox{1}{
    \begin{tabular}{lr|cc|ccccccc}
        \multirow{2}{*}{Model} & \#~Pre-train & \multicolumn{2}{c|}{SNLI-VE} & \multicolumn{6}{c}{Flickr30K~(1K test set)} \\
        &~~Images& dev & test &  IR@1  & IR@5  & IR@10 & TR@1 & TR@5 & TR@10 & RSUM\\
        \shline
        \multicolumn{11}{l}{ \demph{ \it{Pre-trained on More Data} } } \\
        \hline
        \demph{ALIGN\modelbase{}~\citep{jia2021scaling}} & \demph{1.8B} & \demph{-} & \demph{-} & \demph{84.9} & \demph{97.4} & \demph{98.6} & \demph{95.3} & \demph{99.8} & \demph{100.0} & \demph{576.0} \\
        \demph{ALBEF\modelbase{}~\citep{li2021align}} & \demph{14M} & \demph{80.80} & \demph{80.91} & \demph{85.6} & \demph{97.5} & \demph{98.9} & \demph{95.9} & \demph{99.8} & \demph{100.0} & \demph{577.7} \\
        \hline
        \multicolumn{11}{l}{ { \it{Pre-trained on CC, SBU, MSCOCO and VG datasets} } }\\
        \hline
        ViLT\modelbase{}~\citep{kim2021vilt} & 4M & - & - & 64.4 & 88.7 & 93.8 & 83.5 & 96.7 & 98.6 & 525.7 \\
        UNITER\modellarge{}~\citep{chen2020uniter} & 4M & 79.30 & 79.38 & 75.6 & 94.1 & 96.8 & 87.3 & 98.0 & 99.2 & 550.9 \\
        VILLA\modellarge{}~\citep{gan2020large} & 4M & 80.18 & 80.02 & 76.3 & 94.2 & 96.8 & 87.9 & 97.5 & 98.8 & 551.5 \\
        UNIMO\modellarge{}~\citep{li2021unimo} & 4M & \textbf{81.11} & {80.63} & 78.0 & 94.2 & 97.1 & 89.4 & 98.9 & {99.8} & 557.5 \\
        ALBEF\modelbase{}~\citep{li2021align} & 4M & 80.14 & 80.30 & {82.8} & {96.7} & {98.4} & {94.3} & {99.4} &{99.8} & 571.4\\
        \metername{}-CLIP-ViT\modelbase{}~\citep{dou2021meter} & 4M & 80.86 & \textbf{81.19} & {82.2} &  {96.3} & {98.4} & {94.3} & \textbf{99.6} & {99.9} & 570.7 \\
        \methodname{}\modelbase{} (\textbf{Ours}) & \textbf{4M} & \textbf{81.11} & \textbf{81.19} & \textbf{85.8} & \textbf{97.6} & \textbf{98.9} & \textbf{94.7} & \textbf{99.6} & \textbf{100.0} & \textbf{576.6} \\
    \end{tabular}
    }
    \caption{
    Comparisons with previous models on visual entailment (SNLI-VE), image retrieval~(IR) and text retrieval~(TR) tasks (Flickr30K). The best score is bolded.
    }
    \label{tab:snlive_flickr30k_results}
  \end{table*}

\paragraph{Main Results.}

Table~\ref{tab:vqav2_results} and~\ref{tab:snlive_flickr30k_results} show the performance of \methodname{} compared with previous works on downstream VL tasks. 
With only $4$M images for pre-training, \methodname{}\modelbase{} achieves state-of-the-art performance, 
in particular $78.73\%$ accuracy on the VQAv2 test-std set, outperforming the previous state-of-the-art model \metername{} by $1.09\%$ with the same pre-training setting and almost negligible additional parameters and computational costs.
Remarkably, \methodname{}\modelbase{} not only outperforms all base-size models that use the same or a larger number of pre-trained images, but it even outperforms some large-size models.
A similar trend also occurs on the visual entailment and image-text retrieval tasks. 
On the Flickr30K dataset, \methodname{}\modelbase{} achieves the best performance, surpassing not only ALBEF with its specially designed pre-training objective, but also ALIGN with 1.8B pre-train images.

\paragraph{Scaling the Model.}
To investigate the effect of the scale of the model structure on performance, we replace our uni-modal encoders with the corresponding large version, \ie{} RoBERTa\modellarge{} with $355$M parameters and CLIP-ViT-L/14 with $304$M parameters. 
For each layer of the cross-modal encoder, the hidden size is set to $1,024$, the intermediate size of feed-forward networks is set to $4,096$, and the number of heads is set to $16$. The number of cross-modal encoder layers remains $6$ so the number of parameters grows to $200$M.
The patch size is $14 \times 14$, then we set the image resolution to $294 \times 294$ in pre-training and to $574 \times 574$ in fine-tuning on the VQAv2. Other hyperparameters remain unchanged.
As shown in Table~\ref{tab:vqav2_results}, \methodname{} outperforms previous models trained with $10$ times or even $1,000$ times more images, not only in the base size but also in the large size. Notably, \methodname{}\modellarge{} achieves $81.15\%$ accuracy on the VQAv2 test-std set, surpassing the previous state-of-the-art OFA\modellarge{} model by $0.48\%$.
This further demonstrates the effectiveness and scalability of \methodname{}.
In addition, question-answer pairs from VG dataset are often used to extend the VQAv2 training data, thus further improving performance~\citep{teney2018tips, Yu_2019_CVPR}.
Our performance of base and large size can be improved to $79.04\%$ and $81.49\%$ on the VQAv2 test-std set, respectively.

\subsection{Visualization}
Attention mechanism~\citep{bahdanau2014neural} is a critical and naturally interpretable component of transformer-based models. It is intuitive to analyze attention weights since it measures how much attention each token pays to the other tokens.
Inspired by~\citet{xie2022revealing}, we compare the pre-trained \metername{} and \methodname{} models by analyzing the Kullback-Leibler (KL) divergence between attention weight distributions of different attention heads in each layer \footnote{We also follow~\citet{xie2022revealing} to analyze the averaged attention distance and entropy of attention weight distribution between the two pre-trained models, but no significant trends are found.}.
KL divergence can be seen as the diversity of attention heads.
Higher/lower KL divergence means that different attention heads pay attention to different/similar tokens.

\begin{figure}[!t]
  \centering
  \includegraphics[width=0.47\textwidth]{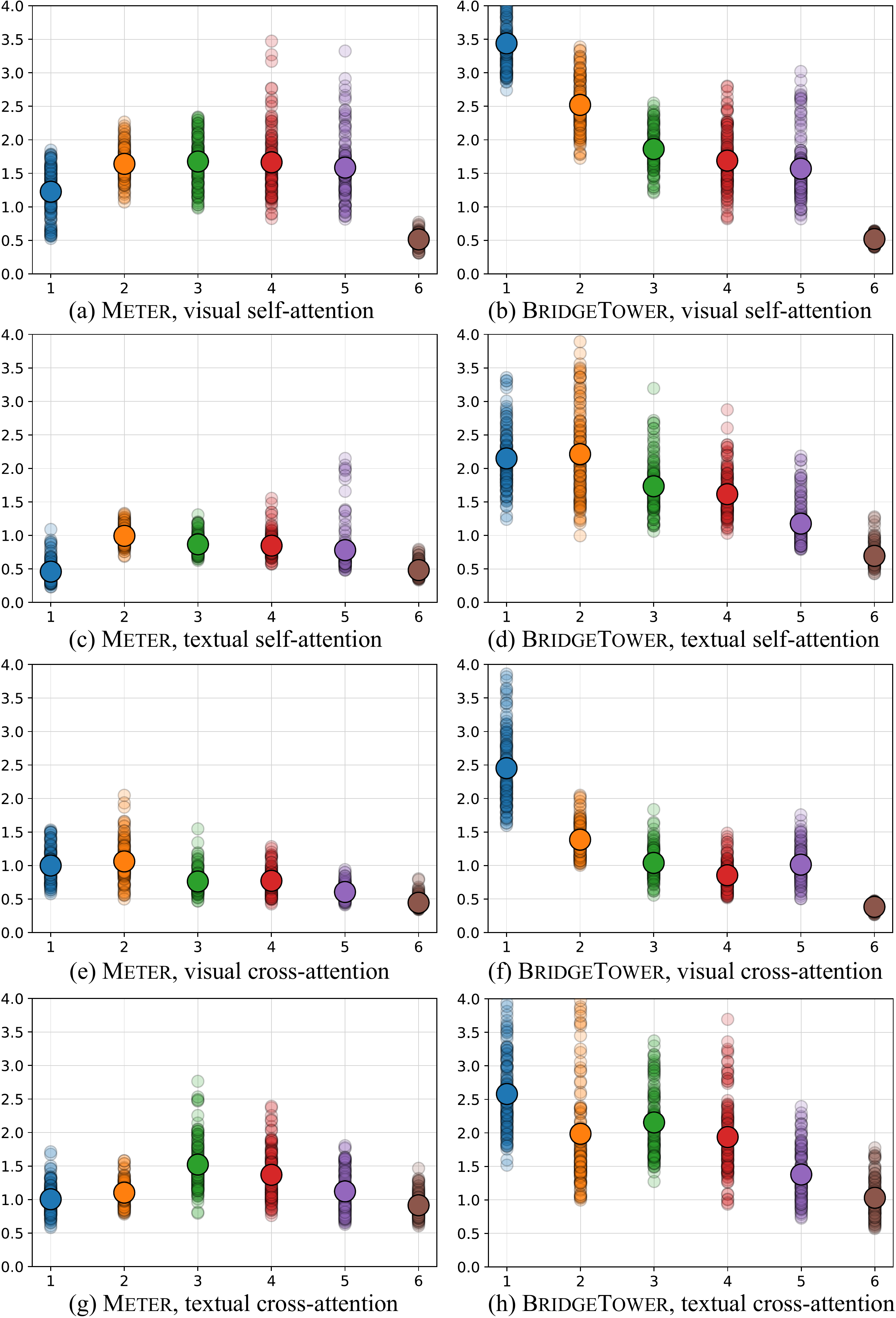}
  \caption{
    The KL divergence between attention distributions of different heads (small dots) and the averaged KL divergence (large dots) in each layer w.r.t. the layer number on the self-/cross-attention of the visual/textual part of the cross-modal encoder in the \metername{} and \methodname{} models.
  }
  \label{fig:KL}
\end{figure}

As shown in Figure~\ref{fig:KL}, by comparing the KL divergence of the two models in each row, there are two distinct trends:
($i$) the diversity of attention heads becomes progressively smaller as the layer goes deeper for \methodname{}, but for \metername{}, the diversity of attention heads becomes progressively larger and then smaller as the layer goes deeper;
($ii$) the diversity of attention heads of each layer of \methodname{} is significantly larger than that of \metername{}, especially for the $1$st to the $5$th layer.
Thus, for different attention heads of self-/cross-attention of the visual/textual part of the cross-modal encoder, compared with \metername{}, \methodname{} can aggregate more different tokens.
We attribute this to our proposed bridge layer, which connects the top layers of uni-modal encoders with each layer of the cross-modal encoder.
Different semantic levels of visual and textual representations are introduced by bridge layers, facilitating more effective and informative cross-modal alignment and fusion at each layer of the cross-modal encoder.

\section{Conclusion and Future Work}
\label{sec:conclusion}
We present~\methodname{}, a simple yet effective vision-language model that introduces multiple bridge layers to build a connection between the top layers of uni-modal encoders and each layer of the cross-modal encoder.
This facilitates effective bottom-up cross-modal alignment and fusion between visual and textual representations of different semantic levels of the pre-trained uni-modal encoders in the cross-modal encoder.
We experimentally prove the effectiveness of the proposed bridge layers and \methodname{}, which achieves remarkable performance in all downstream VL tasks with almost negligible additional parameters and computational costs.
We hope that our work will draw more attention to the rich semantic knowledge latent in the different layers of uni-modal encoders.
Incorporating such semantic knowledge into cross-modal alignment and fusion can yield more expressive and powerful vision-language representations.
Furthermore, experiments with different visual and textual backbones as pre-trained uni-modal encoders demonstrate that the performances of our proposed \methodname{} architecture are consistently and significantly better than that of \metername{}.

In the future, we plan to improve \methodname{} in the following aspects:
\paragraph{Different Pre-training Objectives.}
We followed \metername{} to directly adopt the masked language modeling (MLM) and image-text matching (ITM) as pre-training objectives for a fair comparison. More pre-training objectives, such as image-text contrastive learning (ITC) and masked image modeling (MIM), could be incorporated to investigate their impact on \methodname{} and further improve the performance.
\paragraph{Larger Scale Pre-training.} 
We have pre-trained our model with $4$M images both on the $\text{BASE}$ and $\text{LARGE}$ sizes. 
In both versions, \methodname{} achieves lower accuracy on the ``Number'' type questions of VQAv2 than other models pre-trained with more data. 
We expect to investigate and further improve the performance of \methodname{} after pre-training on larger-scale image-text data.
\paragraph{Generative Task.}
In this paper, we focus on discriminative tasks. 
It would be interesting to investigate the impact of the proposed bridge layer on the performance of a visual language generation task, such as image captioning.

\section*{Acknowledgements}
This work was supported by the National Key R\&D Program of China via grant 2020AAA0106501 and the National Natural Science Foundation of China (NSFC) via grant 62236004 and 61976072.

\appendix


\section{Compare \methodname{} and \metername{}}
\label{appendix:comparision}

In Section 4.2, we use different numbers of cross-modal layers and different visual/textual encoders in the \methodname{} and \metername{} architectures.
The fair comparison with the same pre-trained uni-modal encoders and fine-tune settings on two datasets further demonstrates the pure effect of the proposed \methodname{} architecture. 
Furthermore, after vision-language pre-training (VLP) using the same pre-training data and pre-training settings (Section 4.3), \methodname{} still achieves significant performance improvements over \metername{} on the downstream VL tasks.
An interesting question naturally arises: 
\textit{Will performance improve if we combine the \methodname{} and \metername{} architectures?}

Since the cross-modal layer in the \metername{} architecture is on top of the uni-modal encoders, we denote it as the {external} cross-modal layer, while we denote the cross-modal layer in \methodname{} as the {internal} cross-modal layer.
We fix the number of cross-modal layers as $L_Z=6$ and study the performance of various combinations of {internal} and {external} cross-modal layer numbers.

Table~\ref{tab:design-choice-3} shows that more {internal} layers is more effective than more {external} layers.
This demonstrates that the bridge layer of \methodname{}, which connects the top layers of uni-modal encoders with each layer of the cross-modal encoder, can significantly improve performance. 
Comprehensive and effective bottom-up interactions between uni-modal representations of different semantic levels can be achieved, promoting more effective cross-modal alignment and fusion at each layer of the cross-modal encoder.

We also explore the simple multi-layer feature fusion~\citep{dou2021meter} (last row), where the weighted sum of uni-modal representations of all layers is fed to the top cross-modal encoder. 
Although it obtains a slightly better performance than the $6$ {external} cross-modal layers (penultimate row, \ie{} \metername{}), it is still significantly weaker than the proposed $6$ {internal} cross-modal layers (first row, \ie{} \methodname{}).

\begin{figure*}[!t]
  \centering
  \includegraphics[width=\textwidth]{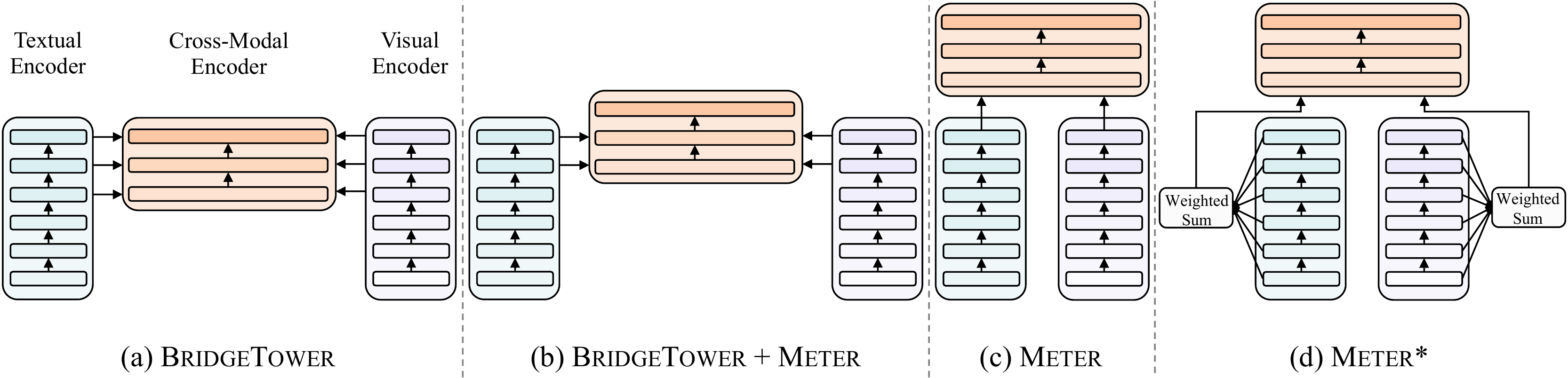}
  \caption{
    (a) \& (b) give brief illustration of \methodname{} and \metername{} architectures; (c) gives a brief illustration of a $3$-layer cross-modal encoder with $2$ internal cross-modal layers and $1$ external cross-modal layer; (d) gives a brief illustration of how \metername{} explores simple multi-layer feature fusion via weighted sum of uni-modal representations of all layers.
  }
  \label{fig:comparision}
\end{figure*}

Figure~\ref{fig:comparision} gives a brief illustration of the four types of architecture mentioned in this section.

\begin{table}[!t]
  \tablestyle{5pt}{1.0}
  \scalebox{1}{
      \begin{tabular}{ccc|l|l}
          \multirow{2}{*}{Type} & \multirow{2}{*}{\#~Internal} & \multirow{2}{*}{\#~External} & \multicolumn{1}{c|}{VQAv2} &  \multicolumn{1}{c}{Flickr30K} \\
          & & & \multicolumn{1}{c|}{Test-Dev} & \multicolumn{1}{c}{RSUM} \\
          \shline
          (a) & 6 & 0 & \textbf{75.18} & \textbf{533.8} \\
          (b) & 4 & 2 & 75.06 ($\downarrow$ 0.12) & 533.1 ($\downarrow$ 0.7) \\
          (b) & 3 & 3 & 74.97 ($\downarrow$ 0.21) & 532.8 ($\downarrow$ 1.0) \\
          (b) & 2 & 4 & 74.71 ($\downarrow$ 0.47) & 532.3 ($\downarrow$ 1.5) \\
          (c) & 0 & 6 & 74.04 ($\downarrow$ 1.14) & 530.7 ($\downarrow$ 3.1)\\
          (d) & 0 & \ \ 6* & 74.52 ($\downarrow$ 0.66) & 531.4 ($\downarrow$ 2.4) \\
      \end{tabular}
  }
  \caption{
    Experiments with different number of {internal} and {external} cross-modal layers.~* denotes simple multi-layer feature fusion. Each type corresponds to a subfigure of the same name in Figure~\ref{fig:comparision}. RSUM indicates the sum of recall metrics for image-text retrieval.
  }
  \label{tab:design-choice-3}
\end{table}

\section{Uni-Modal Tasks}

\begin{table*}[h]
  \tablestyle{5pt}{1.0}
  \scalebox{1.0}{
      \begin{tabular}{l|cc|l}
          Model & CIFAR-10 & CIFAR-100 & \multicolumn{1}{c}{AVG} \\
          \shline
          CLIP-ViT-B-224/16 & 98.74 & 90.64 & 94.69 \\
          + \methodname{} Pre-training & 98.48 & 90.20 & 94.34 ($\downarrow$ 0.35) \\
          + \metername{} Pre-training & 98.46 & 89.52 & 93.99 ($\downarrow$ 0.70) \\
      \end{tabular}
  }
  \caption{
    Linear probe performance of CLIP-ViT-B-224/16 on CIFAR-10 and CIFAR-100 before and after VLP.
  }
  \label{tab:cifar_results}
\end{table*}

\begin{table*}[!h]
  \tablestyle{2.8pt}{1.0}
  \scalebox{1.0}{
    \begin{tabular}{l|cccccccc|l}
      {Model} & MNLI & QQP & QNLI & SST-2 & CoLA & STS-B & MRPC & RTE & \multicolumn{1}{c}{AVG} \\
      & 392k & 363k & 108k & 67k & 8.5k & 5.7k & 3.5k & 2.5k & \\
      \shline
      RoBERTa\modelbase{} & 87.55$\pm$0.03 & 89.22$\pm$0.03 & 92.87$\pm$0.19 & 95.07$\pm$0.30 & 63.21$\pm$2.68 & 90.70$\pm$0.01 & 92.89$\pm$0.35 & 79.90$\pm$0.75 & 86.43 \\
      + BT PT & 87.27$\pm$0.02 & 89.21$\pm$0.06 & 92.78$\pm$0.11 & 94.95$\pm$0.11 & 61.79$\pm$0.83 & 90.85$\pm$0.05 & 92.56$\pm$0.61 & 79.78$\pm$0.96 & 86.15 ($\downarrow$ 0.28)\\
      + \metername{} PT & 87.25$\pm$0.09 & 89.09$\pm$0.07 & 92.72$\pm$0.20 & 94.65$\pm$0.46 & 61.14$\pm$0.45 & 90.65$\pm$0.04 & 92.12$\pm$0.15 & 79.06$\pm$1.08 & 85.84 ($\downarrow$ 0.59) \\
    \end{tabular}
  }
  \caption{
    Fine-tuning performance of RoBERTa\modelbase{} on GLUE dev sets before and after VLP. 
    The number below each task denotes the number of training examples.
    BT is short for \methodname{}. 
    PT is short for Pre-Training.
    We report average scores and standard deviations over three runs of different random seeds.
    Matthews correlations are reported for CoLA, F1 scores are reported for QQP and MRPC, and Spearman correlations are reported for STS-B. 
    The average of matched and mismatched accuracy scores are reported for MNLI.
  }
  \label{tab:glue_results}
\end{table*}

Following previous work~\citep{tan2020vokenization, li2021unimo, dou2021meter}, we report the performance of our visual encoder and textual encoder after VLP on uni-modal tasks, to investigate the effect of VLP on uni-modal encoders of \methodname{}.

For vision tasks, we append a linear classifier to the top of our visual encoder and report the linear probe performance on CIFAR-10 and CIFAR-100~\citep{krizhevsky2009learning}. We perform the grid search over the learning rates and batch sizes for all three models, and the experiments are running on the NVIDIA A100 GPUs.
Table~\ref{tab:cifar_results} shows that, after VLP, the performance of our visual encoder drops slightly on both tasks, but still achieves higher performance compared to \metername{}, especially on CIFAR-100 ($0.68$\% accuracy improvement).
It further proves that the bridge layers of our \methodname{} can improve cross-modal interactions while having slighter impediment \metername{} to the intra-modal interactions of the visual encoder compared to \metername{}.

Recently, LiT~\citep{zhai2021lit} empirically finds that locked pre-trained image models with unlocked text models work best for vision tasks. They believe that image-text data can be great for learning cross-modal alignment between vision and language, but it may not be precise and clean enough to result in state-of-the-art image encoders.
OFA~\citep{wang2022OFA} alleviates this problem by adding the image infilling as a pre-training objective, and achieves competitive fine-tuning performance on ImageNet-1K~\citep{russakovsky2015imagenet}. We will explore the image infilling or masked image modeling~\citep{wang2022image} as a new pre-training objective.

For language tasks, we perform the same grid search as the original RoBERTa\modelbase{}~\citep{liu2019roberta} over the learning rates and batch sizes on the GLUE~\citep{wang2018glue} benchmark. 
All experiments are running on the NVIDIA V100 GPUs.
The best hyperparameters are selected based on the performance over three runs of different random seeds. 
As shown in Table~\ref{tab:glue_results}, after VLP, the performance of our textual encoder degrades slightly on most of the GLUE tasks (except for the STS-B task). 
Similar to the results for the vision tasks, our model outperforms \metername{} on all the GLUE tasks.
This demonstrates that our \methodname{} almost does not affect intra-modal interactions of the textual encoder.

Besides, the same trend is also observed in many current vision-language models~\citep{tan2020vokenization,dou2021meter}, where the performance of the textual encoder decreases slightly after VLP. 
\citet{tan2020vokenization} shows that the texts in image-caption datasets prefer short and instructive descriptions, and thus cause different distributions of sentence lengths and active words compared to the texts in general text-only corpus, \eg{} English Wikipedia and BookCorpus~\citep{zhu2015aligning}. This may contribute to the performance degradation after VLP. Jointly pre-training on the text-only corpus may alleviate this problem~\citep{li2021unimo}.

Notably, in both tasks, we only remain the visual or textual encoder in our \methodname{}\modelbase{} model and ignore the cross-modal encoder. Recent works~\citep{hu2021unit, singh2021flava, sileo2021visual, wang2022OFA} have explored how to utilize the whole vision-language model for uni-modal tasks. We leave this problem as a future direction.

Overall, \methodname{} can better maintain the capability of uni-modal encoders on the CIFAR and GLUE benchmarks compared to \metername{}.
Furthermore, it is a promising direction to explore more uni-modal pre-training objectives.

\section{Inference Time}

\begin{table*}[t]
  \tablestyle{5pt}{1.0}
  \scalebox{1.0}{
    \begin{tabular}{lccc|l|l}
    \multirow{2}{*}{Model} & {\#~Params} & {\#~FLOPs}  & {Time}  & \multicolumn{1}{c|}{VQAv2} &  \multicolumn{1}{c}{Flickr30K}\\ 
    & \multicolumn{1}{r}{(M)} & \multicolumn{1}{r}{(G)} & \multicolumn{1}{c|}{\ (ms)} & \multicolumn{1}{c|}{Test-Dev} & \multicolumn{1}{c}{RSUM} \\
      \shline
        \metername{}-CLIP-ViT\modelbase{} w/o fusion * & 326.56 & \ \ 99.38 & 43.27$\pm$1.18 & 74.04 & 530.7 \\ 
        \metername{}-CLIP-ViT\modelbase{} w/ fusion * & 327.74 & 103.47 & 44.98$\pm$0.86 & 74.52 ($\uparrow$ 0.48) & 531.4 ($\uparrow$ 0.7) \\ 
        \methodname{}\modelbase{} (\textbf{Ours}) & 326.58 & 101.25 & 43.70$\pm$0.99 & 75.18 ($\uparrow$ 1.14) & 533.8 ($\uparrow$ 3.1) \\
    \end{tabular}
  }
  \caption{
      Inference time (ms) and VQAv2 test-dev performance of \metername{} and \methodname{}. * denotes our reimplementation.
  }
  \label{tab:inference_time}
\end{table*}

\begin{table*}[ht]
  \tablestyle{4.7pt}{1.0}
  
  \scalebox{1}{
    \begin{tabular}{lr|cc|ccccccc}
      \multirow{2}{*}{Model} & \#~Pre-train & \multicolumn{2}{c|}{NLVR$^2$} & \multicolumn{6}{c}{MSCOCO~(5K test set)} \\
      &~~Images& dev & test &  IR@1  & IR@5  & IR@10 & TR@1 & TR@5 & TR@10 & RSUM\\
      \shline
      \multicolumn{11}{l}{ \demph{ \it{Pre-trained on More Data} } } \\
      \hline
      \demph{ALIGN\modelbase{}~\citep{jia2021scaling}} & \demph{1.8B} & \demph{-} & \demph{-} & \demph{59.9} & \demph{83.3} & \demph{89.8} & \demph{77.0} & \demph{93.5} & \demph{96.9} & \demph{500.4} \\
      \demph{ALBEF\modelbase{}~\citep{li2021align}} & \demph{14M} & \demph{82.55} & \demph{83.14} & \demph{60.7} & \demph{84.3} & \demph{90.5} & \demph{77.6} & \demph{94.3} & \demph{97.2} & \demph{504.6} \\
      \hline
      \multicolumn{11}{l}{ { \it{Pre-trained on CC, SBU, MSCOCO and VG datasets} } }\\
      \hline
      ViLT\modelbase{}~\citep{kim2021vilt} & 4M & - & - & 42.7 & 72.9 & 83.1 & 61.5 & 86.3 & 92.7 & 439.2 \\
      UNITER\modellarge{}~\citep{chen2020uniter} & 4M & 79.12 & 79.98 & 52.9 & 79.9 & 88.0 & 65.7 & 88.6 & 93.8 & 468.9 \\
      ALBEF\modelbase{}~\citep{li2021align} & 4M & 80.24 & 80.50 & 56.8 & 81.5 & 89.2 & 73.1 & 91.4 & 96.0 & 488.0\\
      \metername{}-CLIP-ViT\modelbase{}~\citep{dou2021meter} & 4M & \textbf{82.33} & {83.05} & {57.1} & {82.7} & {90.1} & \textbf{76.2} & \textbf{93.2} & \textbf{96.8} & 496.1 \\
      \methodname{}\modelbase{} (\textbf{Ours}) & {4M} & {81.85} & \textbf{83.09} & \textbf{62.4} & \textbf{85.1} & \textbf{91.3} & {75.0} & {90.2} & {94.9} & \textbf{498.9} \\
    \end{tabular}
  }
  \caption{
    Comparisons with previous models on visual reasoning (NLVR$^2$), image retrieval~(IR) and text retrieval~(TR) tasks (MSCOCO). The best score is bolded.
  }
  \label{tab:nlvr2_mscoco_results}
\end{table*}

Table~\ref{tab:inference_time} shows the number of parameters, the number
of floating-point operations (FLOPs)\footnote{We use Facebook Research's fvcore library to calculate FLOPs.}, the inference time and downstream performance of \metername{} and \methodname{} without VLP. 
We measure the average inference time of processing 1 VQA instance over 10K runs on 1 NVIDIA Tesla V100-PCIE-32GB GPU. The sequence length is $50$ and the image resolution is $384 \times 384$.
With almost negligible additional parameters (18.4K parameters), inference time (less than $0.5$ ms), and computational costs ($\frac{101.25}{99.38}\approx 1.0188$), \methodname{} achieves $1.14\%$ and $3.1\%$ performance improvement over \metername{} on the VQAv2 test-dev score and Flickr30K RSUM score.
This demonstrates the efficiency and effectiveness of \methodname{} compared with \metername{}.

\begin{figure*}[t]
  \centering
  \includegraphics[width=0.8\textwidth]{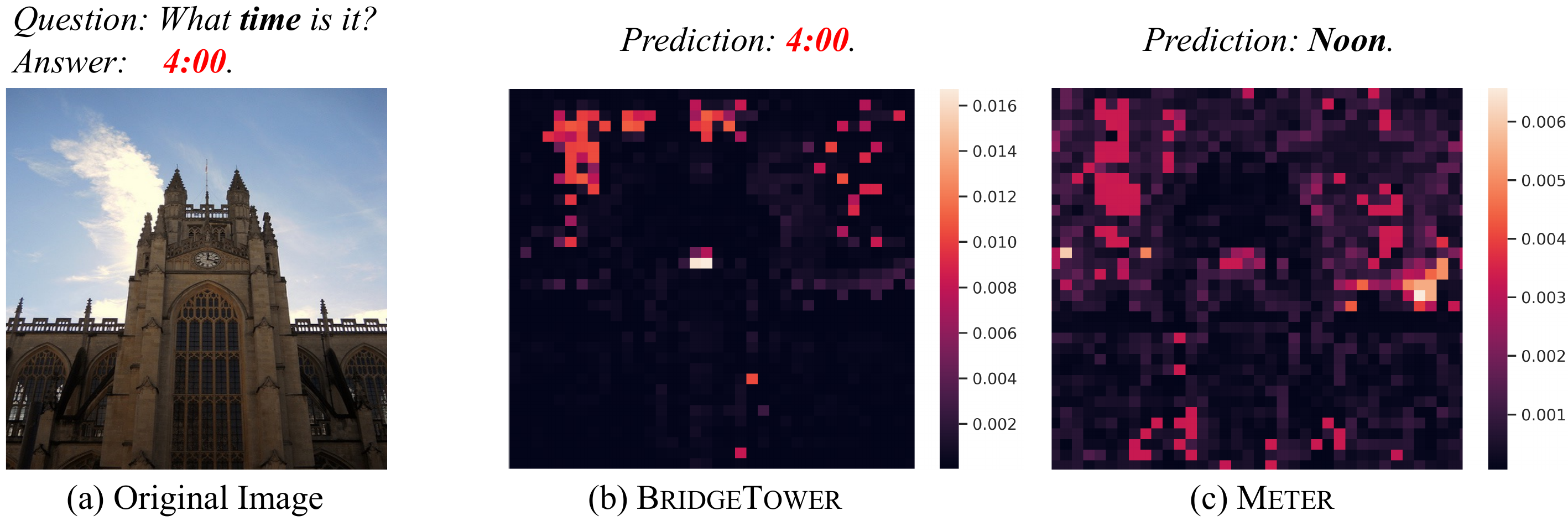}
  \caption{
      Visualization of the cross-attention map of our \methodname{} and \metername{}. The example comes from the VQAv2 validation set. Predictions come from the fine-tuning checkpoints of both models.
  }
  \label{fig:visualization}
\end{figure*}

\section{Performance on Other Downstream Tasks}
We also evaluate \methodname{} on visual reasoning~\citep{suhr2019corpus} (NLVR$^2$), and image-text retrieval~\citep{lin2014microsoft} (MSCOCO) tasks.
We fine-tune our \methodname{} with the strategy in Sec. 3.5.
For MSCOCO dataset, we follow the standard Karpathy Split~\citep{karpathy2015deep}.
Table~\ref{tab:nlvr2_mscoco_results} shows that, compared with previous work, \methodname{} achieves either best or competitive performance on the NLVR$^2$ and MSCOCO dataset. 
In the MSCOCO dataset, \methodname{} achieves an RSUM of $498.9\%$, outperforming the previous state-of-the-art model \metername{} by $2.8\%$ with the same pre-training setting and almost negligible additional parameters and computational costs.
In particular, for the image retrieval task on the MSCOCO dataset, \methodname{} achieves $62.4\%$ for IR@1, which not only significantly surpasses \metername{} by $5.3\%$, but also surpasses the ALIGN and ALBEF models pre-trained with orders-of-magnitude larger datasets. For the text retrieval task on the MSCOCO dataset, \methodname{} achieves $75.0\%$ for TR@1, which is lower than \metername{} by $1.2\%$. We leave the reason for the different performance of image retrieval and text retrieval tasks on the MSCOCO dataset for future study.

\section{Detailed VQAv2 Performance Comparison}

\begin{table*}[t]
    \tablestyle{5pt}{1.0} 
    \scalebox{1.0}{
      \begin{tabular}{lccc|c}
        \multirow{2}{*}{Model} & {Image Size} & {\#~Pre-train} & {Use VG--QA} & {Test-Dev} \\
          & For Pre-training & Images & For Fine-tuning & Overall \\
        \shline
      \metername{}-CLIP-ViT\modelbase{} w/o fusion & - & - & \xmark & 71.75 \\
      \metername{}-CLIP-ViT\modelbase{} w/ fusion& - & - & \xmark & 72.92 \\
      \metername{}-CLIP-ViT\modelbase{} w/o fusion & 224 & 4M & \xmark & 77.19 \\
      \metername{}-CLIP-ViT\modelbase{} w/ fusion & 224 & 4M & \xmark & 77.06 \\
      \hline
      \metername{}-CLIP-ViT\modelbase{} w/o fusion * & - & - & \xmark & 74.04 \\
      \metername{}-CLIP-ViT\modelbase{} w/ fusion * & - & - & \xmark & 74.52 \\
      \methodname{}\modelbase{}  & - & - & \xmark & 75.18 \\
      \hline
      \metername{}-CLIP-ViT\modelbase{} & 288 & 4M & \xmark & 77.68 \\
      \methodname{}\modelbase{}  & 288 & {4M} & \xmark & {78.66} \\
      \methodname{}\modelbase{}  & 288 & {4M} & \cmark(COCO-only) & {78.77}  \\
      \methodname{}\modelbase{}  & 288 & {4M} & \cmark & {79.10}  \\
      \hline
      \metername{}-CoSwin\modelhuge{} & 384 & 14M~~ & \xmark & 80.33\\
      \methodname{}\modellarge{} & 294 & {4M} & \xmark & {81.25} \\
      \methodname{}\modellarge{} & 294 & {4M} & \cmark(COCO-only) & {81.32} \\
      \methodname{}\modellarge{} & 294 & {4M} & \cmark & {81.52} \\
    \end{tabular}
    }
    \caption{
      Detailed performance comparisons between \metername{} and \methodname{} on visual question answering (VQAv2) both with and without pre-training. The results in the first block come from \metername{}~\citep{dou2021meter}. * denotes our re-implementation. 
    }
    \label{tab:vqav2_detailed}
  \end{table*}

Table~\ref{tab:vqav2_detailed} shows the detailed performance comparison between \metername{} and \methodname{} on VQAv2.
All experiments use an image resolution of $576 \times 576$ for fine-tuning.

\paragraph{Comparison Between \metername{} and \methodname{}.}
In the first block, \metername{} shows that the fusion strategy can improve performance by a small margin without pre-training, but it can degrade performance after pre-training.
We reimplemented the \metername{} model with and without fusion strategy in the second block, which has been shown in Section 4.2.3 of the paper.
We perform grid searches over the learning rates, which may contribute to the better performance ($71.75\%$$-$$>$$74.04\%$, $72.92\%$$-$$>$$74.52\%$) compared to the same setting in \metername{}\footnote{Besides, the pre-trained checkpoints released in \metername{}'s github code repository omit the last layer of CLIP-ViT-B-224/16. In the issues/9, they explain that using 11 layers can get slightly better performance (about $0.5\%$ VQA score improvements). However, in our experiments, we find that using all 12 layers achieves better performance (for \metername{} and \methodname{} architectures without VLP), then we use all 12 layers in all base size experiments.}.
Without VLP, our base model achieves an accuracy of $75.18\%$ on the VQAv2 test-dev set, which outperforms \metername{} both with and without fusion strategy.
With VLP, our base model achieves an accuracy of $78.66\%$ on VQAv2, which still surpasses both versions of \metername{}.
This demonstrates that through the connection established by bridge layers, our \methodname{} achieves comprehensive and detailed layer-wise interaction between the uni-modal representations of different semantic levels.

\paragraph{Utilization of VG-QA for Fine-Tuning.}
In the third and fourth blocks, we show the VQAv2 fine-tuning performance of our \methodname{}\modelbase{} model and \methodname{}\modellarge{} model with different settings of addition question-answer pairs from Visual Genome~\citep{krishna2017visual}.
Following the standard practice~\citep{teney2018tips, Yu_2019_CVPR}, we only use question-answer pairs in Visual Genome where the correct answer appears in the answer classes of VQAv2. 
We denote the valid $932$k question-answer pairs in Visual Genome as VG-QA. 
In VG-QA, there are $468$k question-answer pairs using images that are also used in VQAv2, and we refer to this part of the data as VG-QA(COCO-only).
As shown in the last two rows of the third and fourth blocks, VG-QA(COCO-only) can slightly improve the performance of the base model and large model ($0.11\%$ and $0.07\%$, respectively). 
Using the whole VG-QA can improve our performance to $79.10\%$ and $81.52\%$ on the VQAv2 test-dev set in the base and large size, respectively.

\begin{table}[t]
  \tablestyle{5pt}{1.0}
  \scalebox{1.0}{
    \begin{tabular}{l|cccc}
      & COCO & VG & CC & SBU \\
      \shline
      \#~Images & 113K & 108K & 2.9M & 860K \\
      \#~Captions & 567K & 4.8M & 2.9M & 860K \\
    \end{tabular}
  }
  \caption{
    Statistics of the pre-training datasets. We remove duplicate image-caption pairs in VG~\citep{kim2021vilt, dou2021meter} and only 2.9M image-caption pairs can be downloaded in CC.
  }
  \label{tab:statistics_pretrain}
\end{table}

\section{Case Study}

In this section, we use the heatmap of the attention weights to visualize the difference between our \methodname{} and \metername{}.
We visualize the cross-attention map from the selected token ``\textbf{time}'' to the image. We use the average attention weights from all attention heads at the last layer of the cross-modal encoder.
As shown in Figure~\ref{fig:visualization}, given the question: ``\textit{What time is it?}'', our \methodname{} can correctly attend to the clock in the clock tower and the sky in the background and predict the time shown on the clock, while the \metername{} is distracted by the sky in the background.
We attribute this to the fact that \metername{} focuses on the global information in the last-layer uni-modal features, while our \methodname{} can obtain the different semantic levels of visual and textual representations through multiple bridge layers, and interact thoroughly and mildly in the bottom-up directions. Therefore, comprehensive and detailed layer-wise interactions in the cross-model encoder help our \methodname{} to predict the time in the clock.

\section{Experimental Settings}

\begin{table*}[!h]
  \tablestyle{5pt}{1.0}
  \scalebox{1.0}{
    \begin{tabular}{l|cc}
      Hyperparameters & \methodname{}\modelbase{} & \methodname{}\modellarge{} \\
      \shline
      Number of Layers & $6$ & $6$ \\
      Hidden size & $768$ & $1,024$ \\
      FFN inner hidden size & $3,072$ & $4,096$ \\
      Number of Attention heads & $12$ & $16$ \\
      Dropout Ratio & $0.1$ & $0.1$ \\
      Attention dropout & $0.1$ & $0.1$ \\
      \hline
      Total Steps & $100$k & $100$k \\
      Batch Size & $4,096$ & $4,096$ \\
      Optimizer & $\operatorname{AdamW}$ & $\operatorname{AdamW}$ \\
      Learning Rate & $1e^{-5}$ & $1e^{-5}$ \\
      Learning Rate Decay & $\operatorname{Linear}$ & $\operatorname{Linear}$ \\
      Weight Decay & $0.01$ & $0.01$ \\
      Warmup Steps & $10$k & $10$k \\
      Adam $\epsilon$ & $1e^{-8}$ & $1e^{-8}$ \\
      Adam $\beta_1$ & $0.9$ & $0.9$ \\
      Adam $\beta_2$ & $0.98$ & $0.98$ \\
      \hline
      Center-Crop & \cmark & \cmark \\
      Random Resized Crop & \xmark & \xmark \\
      Random Augmentation & \xmark & \xmark \\
      Random Horizontal Flipping & \xmark & \xmark \\
      \hline
      Textual Encoder & RoBERTa\modelbase{} & RoBERTa\modellarge{} \\
      Visual Encoder & CLIP-ViT-B & CLIP-ViT-L \\
      Patch Size & $16$ & $14$ \\
      Image Resolution for VLP & $288$ & $294$ \\
    \end{tabular}
  }
  \caption{
    Hyperparameters for pre-training \methodname{}\modelbase{} and \methodname{}\modellarge{}. The first block is the hyperparameters for the cross-modal encoder in our \methodname{} model.
  }
  \label{tab:hyperparam_pretrain}
\end{table*}

\begin{table*}[!h]
  \tablestyle{5pt}{1.0}
  \scalebox{1.0}{
    \begin{tabular}{l|ccc}
      Hyperparameters & VQAv2 & SNLI-VE & Flickr30K \\
      \shline
      Total Epochs & $10$ & $5$ & $20$ \\
      Batch Size & $512$ & $64$ & $512$ \\
      Optimizer & $\operatorname{AdamW}$ & $\operatorname{AdamW}$ & $\operatorname{AdamW}$ \\
      Learning Rate & $1e^{-5}$~/~$4e^{-6}$ & $3e^{-6}$ & $5e^{-6}$ \\
      Learning Rate Decay & $\operatorname{Linear}$ & $\operatorname{Linear}$ & $\operatorname{Linear}$ \\
      Weight Decay & $0.05$ & $0.01$ & $0.01$ \\
      Warmup Ratio & $0.06$ & $0.06$ & $0.05$ \\
      Adam $\epsilon$ & $1e^{-8}$ & $1e^{-8}$ & $1e^{-8}$ \\
      Adam $\beta_1$ & $0.9$ & $0.9$ & $0.9$ \\
      Adam $\beta_2$ & $0.98$ & $0.98$ & $0.98$ \\
      \hline
      Center-Crop & \xmark & \xmark & \xmark \\
      Random Resized Crop & \cmark & \cmark & \cmark \\
      Random Augmentation & \cmark & \cmark & \cmark \\
      Random Horizontal Flipping & \xmark & \cmark & \cmark \\
      \hline
      Textual Encoder & RoBERTa$_\text{BASE~/~LARGE}$ & RoBERTa\modelbase{} & RoBERTa\modelbase{} \\
      Visual Encoder & CLIP-ViT-B~/~L & CLIP-ViT-B & CLIP-ViT-B \\
      Patch Size & $16$~/~$14$ & $16$ & $16$\\
      Image Resolution for VLP & $288$~/~$294$ & $288$ & $288$ \\
      Image Resolution for FT & $576$~/~$574$ & $384$ & $384$ \\
      Loss Function & BCE & CE & CE \\
    \end{tabular}
  }
  \caption{
    Hyperparameters for fine-tuning \methodname{} on downstream VL tasks. FT denotes fine-tuning. CE and BCE are short for cross-entropy loss and binary cross-entropy loss, respectively. 
  }
  \label{tab:hyperparam_finetune}
\end{table*}

\paragraph{Pre-training Details}
Table~\ref{tab:statistics_pretrain} shows the statistics of our pre-training datasets.
Following previous work~\citep{kim2021vilt,chen2020uniter,li2021align,dou2021meter}, we adopt four public image-caption datasets for pre-training, including Conceptual Captions (CC)~\citep{sharma-etal-2018-conceptual}, SBU Captions (SBU)~\citep{NIPS2011_5dd9db5e}, MSCOCO Captions (COCO)~\citep{chen2015microsoft}, and Visual Genome (VG)~\citep{krishna2017visual}. 
The total numbers of the unique images and image-caption pairs in the combined training data are $4$M and $9$M. Table~\ref{tab:hyperparam_pretrain} describes the hyperparameters for pre-training of our \methodname{}\modelbase{} and \methodname{}\modellarge{} models. 

\paragraph{Hyperparameters for Fine-Tuning Downstream Tasks.}
Similar to the image-text matching (ITM) pre-training objective, we pass the final representation of $\texttt{[class]}$ token and $\texttt{[<s>]}$ token to the non-linear layer activated by $\texttt{Tanh}$, and feed the concatenation of the output into a linear classifier (Flickr30K) or an MLP classifier(VQAv2 and SNLI-VE). We apply cross-entropy loss for SNLI-VE and Flickr30K and binary cross-entropy loss for VQAv2~\citep{kim2021vilt,dou2021meter}. 
Fine-tuning hyperparameters for VQAv2, SNLI-VE, and Flickr30K are given in Table~\ref{tab:hyperparam_finetune}.

\clearpage

\bibliography{aaai23_long}

\end{document}